\documentclass[runningheads]{llncs}
\usepackage{times}  
\usepackage{helvet}  
\usepackage{courier}  
\usepackage{url}  
\usepackage{graphicx}  
\usepackage{amsmath}
\usepackage{float}
\input{epsf}
\begin{document}
\title{TrolleyMod v1.0: An Open-Source Simulation and Data-Collection Platform for Ethical Decision Making in Autonomous Vehicles 
 }

\titlerunning{TrolleyMod v1.0}
%
\author{Vahid Behzadan\inst{1} \and 
James Minton\inst{1} \and
Arslan Munir\inst{1}} 

\authorrunning{V. Behzadan et al.}
%
\institute{Kansas State University, Manhattan, KS 66506, USA\\
\email{\{behzadan,jzm,amunir\}@ksu.edu}\\
\url{http://blogs.k-state.edu/aisecurityresearch/}}
\maketitle              
\begin{abstract}
	This paper presents TrolleyMod v1.0, an open-source platform based on the CARLA simulator for the collection of ethical decision-making data for autonomous vehicles. This platform is designed to facilitate experiments aiming to observe and record human decisions and actions in high-fidelity simulations of ethical dilemmas that occur in the context of driving. Targeting experiments in the class of trolley problems, TrolleyMod provides a seamless approach to creating new experimental settings and environments with the realistic physics-engine and the high-quality graphical capabilities of CARLA and the Unreal Engine. Also, TrolleyMod provides a straightforward interface between the CARLA environment and Python to enable the implementation of custom controllers, such as deep reinforcement learning agents. The results of such experiments can be used for sociological analyses, as well as the training and tuning of value-aligned autonomous vehicles based on social values that are inferred from observations.
\end{abstract}

\section{Introduction}
With the rise in technological advancements and investments in autonomous vehicles, the issue of ethical decision-making in such systems is becoming growingly pronounced \cite{goodall2014machine}. Similar to human drivers, the Artificial Intelligence (AI) controllers of driverless vehicles will inevitably face moral dilemmas, which cannot be solved by adherence to simple ethical principles \cite{dwork2012fairness}. A well-known instance of such dilemmas is the \emph{trolley problem} \cite{jarvis1985trolley}, where by some means (e.g., brake malfunction), the vehicle is put in a situation that forces the driver to decide between hurting bystanders on one side or another. Considering the enhanced computational and observational abilities of AI, it is expected that autonomous vehicles make \emph{better} decisions than human drivers in such circumstances. However, the definition of \emph{better} is deeply rooted in complex ethical principles and policies that are difficult (if not practically impossible) to formally specify \cite{greene2016embedding}. Furthermore, such principles are often dynamic across cultural, geographical, and temporal dimensions \cite{awad2018moral}.

While the problem of ethical decision-making in AI has remained an open challenge for many decades \cite{wallach2008moral}, recent studies on data-driven approaches to this problem have reported promising advances towards practical solutions \cite{kasenberg2018norms}. Another advancement is the proposal to model the dynamics of ethical decision-making within the abstraction of \emph{value-alignment}, which reduces the problem to the identification and measurement of the ethical norms and values of the society, and implementing such values into AI \cite{arnold2017value}. The quantification and modeling of such norms and values can be performed via machine learning and parametric techniques, examples of which include those that are based on Inverse Reinforcement Learning (IRL) \cite{abel2016reinforcement} and norm inference \cite{kasenberg2018norms}. A major hurdle in such approaches is that of data collection \cite{kim2017computational}. Modern machine learning approaches and model-dependent techniques require a large number of samples that provide a representative dataset of the societal choices and ethical values \cite{noothigattu2017voting}. In particular, due to the rarity of representative ethical dilemmas in controlled real-world observations, the majority of current studies resort to simulation-based experiments.

However, the experimental setups and configurations utilized by such studies suffer from a number of shortcomings. Firstly, the bulk of such setups do not provide the means for measuring the performance of human operators in real-time, and thus fail to present a baseline for benchmarking the performance of AI against humans. Secondly, the majority of simulation environments used in published experiments (e.g., \cite{awad2018moral}) provide a very simple depiction of the conditions in real-world dilemmas, and hence may fail to account for salient details that are crucial to ethical decision-making. Thirdly, most of the recent experiments are performed in simulation environments that are either not open-source, or are very difficult to reconfigure and customize for new experiments.

In response to the aforementioned gap, this paper presents TrolleyMod v1.0, an open-source simulation platform based on the CARLA simulator for autonomous driving research \cite{dosovitskiy2017carla}. Targeting experiments in the class of trolley problems, TrolleyMod provides a seamless approach to creating new experimental settings and environments with the realistic physics-engine and the high-quality graphical capabilities of CARLA and the Unreal Engine \cite{games2007unreal}. Also, TrolleyMod provides a straightforward interface between the CARLA environment and Python to enable the implementation of custom controllers, such as deep reinforcement learning agents in Tensorflow (e.g., \cite{liang2018cirl}) or Pytorch (e.g., \cite{genandercontrol}). The details of TrolleyMod are presented in the remainder of this paper, organized as follows: We present an overview of experimental platforms used in data-driven ethical decision-making studies, followed by the architectural details and components of TrolleyMod. We conclude the paper with remarks on plans for future extensions of this project.

\section{Previous Work}\label{Sec:Previous}
While the philosophical debate and research on ethical decision-making in AI is decades old, the engineering work in this area is very recent \cite{kasenberg2018norms}. In particular, the interest in creating moral autonomous agents has rapidly grown in the past few years. Specifically, advances in imitation learning \cite{taylor2016alignment}, inverse game theory \cite{wang2017using}, and IRL \cite{abel2016reinforcement} have triggered a growing investment into the research on data-driven approaches to the modeling and transfer of human ethics to autonomous agents. Such approaches rely on collection of data on ethical decisions and actions of many human subjects to derive a representative model of the societal ethical principles \cite{noothigattu2017voting}. 

In the domain of autonomous vehicles, very few platforms for this purpose are reported, and even fewer are openly available to the research community. Of the best known ethical data collection platforms for autonomous vehicles is MIT's Moral Machine \cite{kim2018computational} project, which performs a large-scale crowdsourcing of ethical opinions for a few simple cases of the trolley problem. While this project has yielded many interesting results \cite{awad2018moral}, it can be argued that the low fidelity of the experiments, as well as the inconsideration of temporal and cultural externalities, diminish the feasibility of inferring practical ethical policies that can be implemented in real-world driverless vehicles. Furthermore, the Moral Machine experiment fails to account for the effect of environmental nudges \cite{leonard2008richard} in the simulated scenarios. Also, while the datasets and analytics of this project were recently made available, the code for the simulation software itself is not accompanied, and hence does not accommodate the adoption, extension, and customization of this platform for further research.

While at a smaller scale, the Ethical Autonomous Vehicles\footnote{\url{http://mchrbn.net/ethical-autonomous-vehicles/}} project also provides the means for data collection and experimentation on ethical decision making. Albeit, this project seems to be outdated due to lack of maintenance. Also, the limited scenarios and flexibility of this project, as well as the lack of documentation, may render the usability of this project for extended research infeasible. 

In \cite{frison2016first}, authors present an alternative data collection approach using a kinetic high-fidelity driving simulator equipped with a fully automated autonomous driver. Unfortunately, the authors do not provide more information on the setup and replication procedures for their experiments.

While these prior projects succeeded in providing very valuable insights into the problem of ethical decision-making, there remains a need for an open-source, well-documented, high-fidelity, and highly-flexible simulation platform to further facilitate experimentation and research on crowdsourced ethical models for autonomous vehicles.

\section{TrolleyMod v1.0}\label{Sec:Trolley}
The goal of TrolleyMod is to provide an easy-to-use and flexible platform for setting up the conditions pertaining to ethical dilemmas. While the current version is mainly focused on the variations of trolley problem, this platform supports seamless customization and adjustment of the experiment to broader classes of ethical decision-making, such as abiding by the traffic laws. TrolleyMod is designed to facilitate experiments that require high-fidelity simulations of ethical dilemmas that occur in driving tasks to observe and record human decisions. The results of such experiments can be used for sociological analyses, as well as the training and tuning of value-aligned AI agents for autonomous driving based on social values that are inferred from observations.
\begin{figure}[h]
	\centering
	\includegraphics[width=\columnwidth]{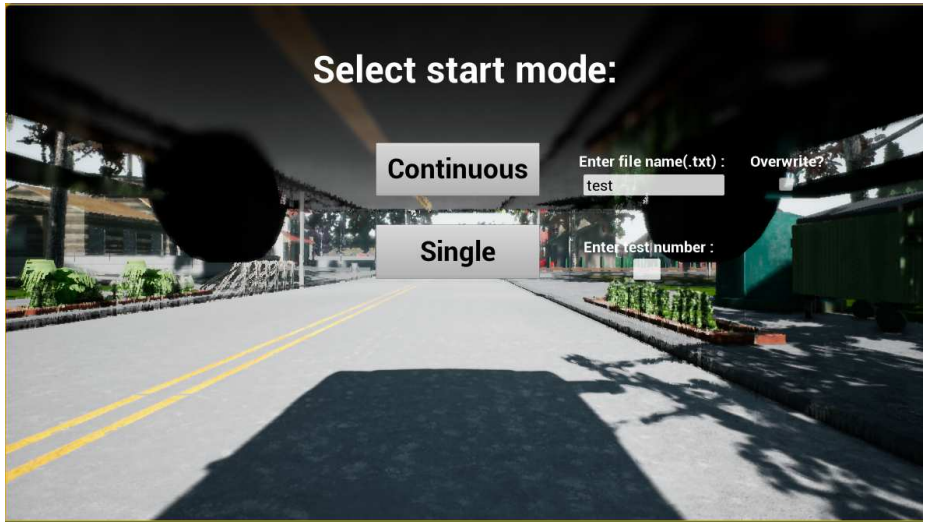}
	\caption{Startup menu of a TrolleyMod simulation}
	\label{Fig:First}
\end{figure}

In TrolleyMod, the trolley problems are set up such that there are one or more victims on each side of a road belonging to a simulation environment. The environment (i.e., map) may be designed to represent realistic conditions in urban, suburban, highway, or custom settings. In each episode of the experiment, the subject is put in the driver's seat of a vehicle that accelerates automatically. The only control actions available to the subject are swerving left or right. Furthermore, the conditions of dilemma can be reinforced via invisible barriers to ensure that the only available options to the driver are to collide with at least one group of victims or other objects. Upon collision, the subject's actions are written into a text file or a network socket for processing. In TrolleyMod, each individual Trolley problem is called a \emph{scenario}. A sequence of scenarios is what constitutes a TrolleyMod simulation. The components and procedures of a simulation are detailed as follows:

\subsection{CARLA}
As the name suggests, TrolleyMod is a modification of the CARLA (Car Learning to Act) simulator \cite{dosovitskiy2017carla}. CARLA is an open-source simulator for urban driving, designed to support training, prototyping, and validation of autonomous driving models. This platform allows for the flexible configuration of sensor suites and provides various signals useful to the training of driving tasks, such as GPS coordinates, speed, acceleration, and detailed data on collisions. CARLA is implemented over Unreal Engine 4 \cite{games2007unreal} to provide flexibility and realism in the physics and high-fidelity visualization of driving environments. CARLA follows a client-server architecture to provide an interface between the world and various types of agents. In this architecture, the server runs the simulation and renders the scene, while the client uses a Python API to interact with the simulation. 

The environments of CARLA are composed of 3D models of static objects, such as traffic signs, buildings, infrastructure, vegetation, as well as dynamic objects, such as vehicles and pedestrians. The default sensors in CARLA are comprised of RGB (Red, Green, and Blue) camera and pseudo-sensors which provide semantic segmentations in terms of road, traffic sign, sidewalk, pedestrian, etc. Furthermore, CARLA provides various measurements associated with the state of the agent as well as compliance with traffic rules. Such measurements include orientation and location, acceleration vector, speed, and the cumulative impact of collisions. Signals corresponding to traffic rules include the state of traffic lights and speed limits. 

\subsection{Components of TrolleyMod}
There are three types of objects that can be spawned automatically with TrolleyMod: pedestrians, vehicles, and prop objects. These correspond to the Walker, the CarlaWheeledVehicle, and StaticActor classes respectively. The first two classes are native to CARLA, while StaticActor is a product of TrolleyMod. Our extension adjusts these classes to enable their instance to store new information pertaining to trolley scenarios. All three classes share two variables:
\begin{enumerate}
	\item TestNum: an integer that tracks what scenario is being run.
	\item GroupMemberNames: a delineated string that lumps together the names and properties of all victims in the scenario to which the class instance belongs.
\end{enumerate}

Walker actors also retain the following additional information for the pedestrian character: age, gender, ID number of the pedestrian's group in the scenario, size of the pedestrian's group, and special traits (e.g., pregnant, disabled, etc.)

\begin{figure*}[t]
	\centering
	\includegraphics[width=\textwidth]{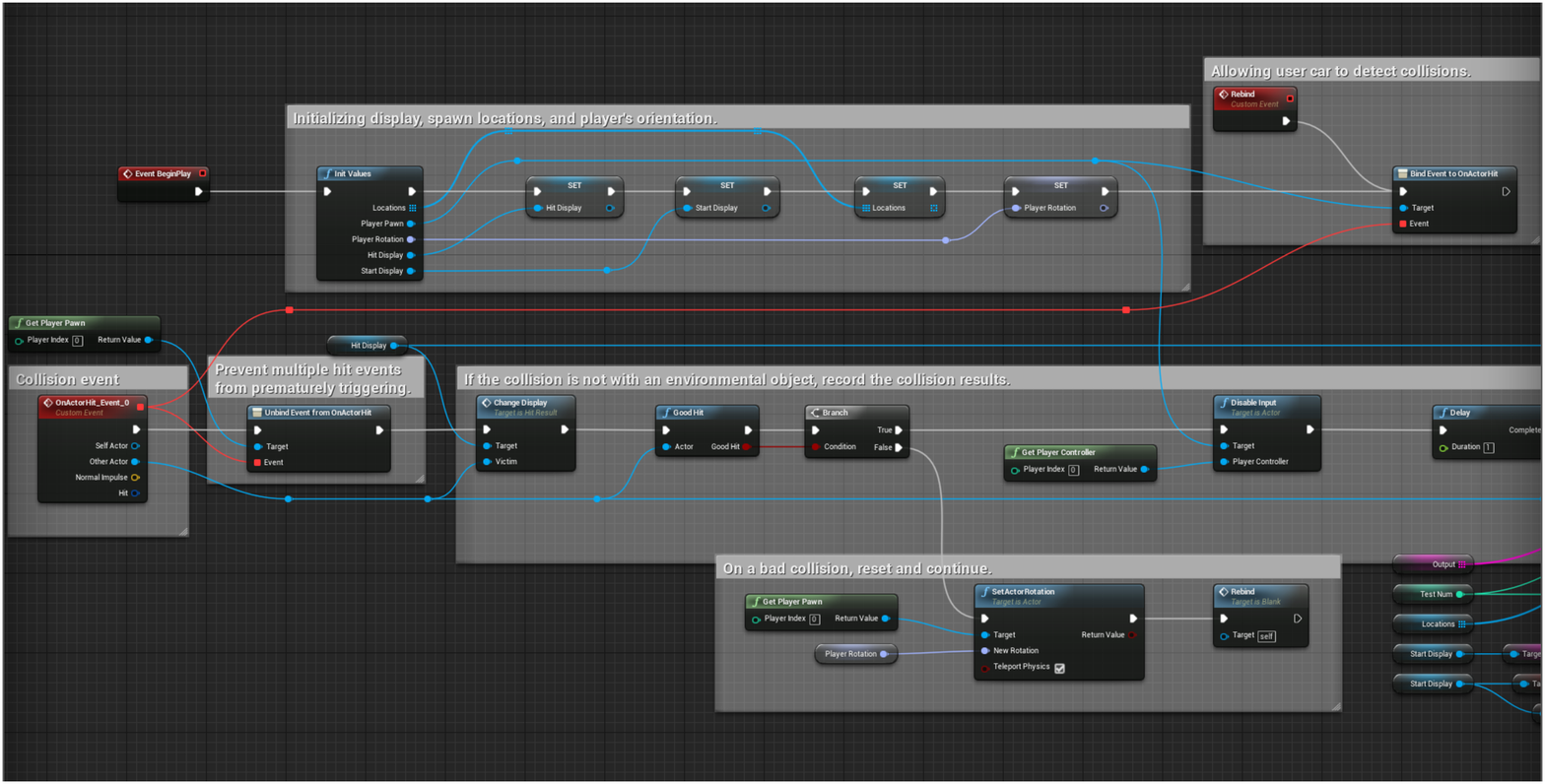}
	\caption{Blueprint functions for execution and control of the simulation}
	\label{Fig:Blueprint1}
\end{figure*}

\subsection{Configuration}
TrolleyMod spawns object models into a ``level'' (i.e., episode of simulation) in the Unreal Engine 4. It takes an input text file, which contains a list of tuples specifying an object (e.g., pedestrian, car, cones, etc.) that can be collided with by the subject and provides any other ancillary data that describes the object. This text is parsed to provide a reference to Unreal Engine's Actor class in memory. TrolleyMod implements a custom format for the specification of objects in a scenario, as detailed in the online documentation\footnote{\url{https://github.com/zminton/TrolleyMod/wiki}}.

\begin{figure}[h]
	\centering
	\includegraphics[width=\columnwidth]{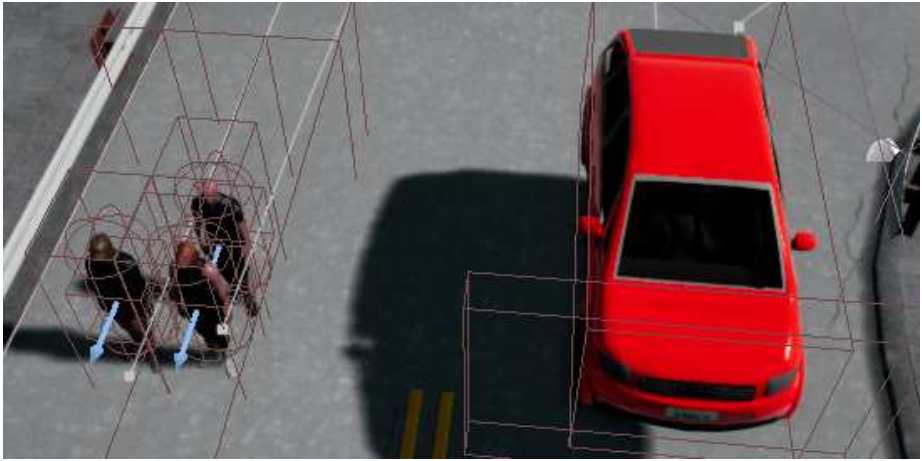}
	\caption{Manual configuration of Actor objects}
	\label{Fig:Config}
\end{figure}

To generate a scenario in TrolleyMod, the experiment designer runs a generation Blutility ((Editor Utility Blueprint)) named ScenarioGen. A blutility is an Unreal Engine object that enables execution of functions outside of the runtime. In TrolleyMod, blutilities are essential for providing a means to quickly modify the CARLA-provided maps to facilitate the generation of various scenarios in a relatively short period of time.

After the ScenarioGen blutility is run, the experiment designer may add other scenarios in unused locations in the map by manually placing Actor objects into the level, as illustrated in Fig.~\ref{Fig:Config}. Also, for the data recording script to recognize a new scenario, a Target object must be placed into the map first. 

TrolleyMod provides the necessary functions and assets to set up scenarios, but it is up to the user on how to arrange and utilize them in customized environments and settings. Since each user will want to define a simulation differently, TrolleyMod only provides a functional framework. Further details on this procedure is available in the online documentation.

\subsection{Execution of Simulation}
TrolleyMod utilizes a modular structure based on \emph{Blueprint Function Library}, which is an Unreal Engine feature that allows for the functions in the library to be reused over multiple projects. In TrolleyMod, blueprints provide a prepackaged set of functions to be used in the level blueprint of maps derived from CARLA's native environments. Currently, TrolleyMod contains only one blueprint, named ``FunctionLibrary''. This blueprint contains all of the functions required to run TrolleyMod simulations. When the subject's agent collides with a victim, that result is recorded and the next scenario is spawned. Once all scenarios are complete, the simulation terminates and the results are written to a text file or network socket.

Each TrolleyMod simulation includes an \emph{Event Tick} node, which is called at every frame during the simulation. The purpose of this node is to make the car controlled by the subject to automatically accelerate, thus forcing the subject to decide quickly on what victim to collide with. As depicted in Fig.~\ref{Fig:Blueprint1}, the initial event object of this blueprint (i.e., \emph{BeginPlay}) first calls the \emph{InitValues} function in order to initialize the environment and object attributes. Then, it binds the subject's agent to a hit event --that is, whenever the agent collides with any object in the map, it will generate an event that calls the \emph{OnActorHit\_Event\_0 node}. Next, assuming the agent hits one of the intended victims, the blueprint generates a text message describing what the player collided with using the \emph{SetDisplay} function. If the player does not hit an intended target, the simulation refers to the \emph{PlayerRotation} variable to reset where the car is facing and rebinds hit events to it.

Otherwise, the \emph{CollisionHandler} function analyzes the Actor hit by the player, extracts its data to the Output variable, and then prepares the simulation to move on to the next scenario. This continues until all the scenarios in the \emph{Locations} array have been visited, after which the simulation writes the data to a text file. It is also possible for the player to specify only one scenario to complete using the start menu, and consequently the simulation will terminate after the specified scenario is run.

\section{Future Work}\label{Sec:Future}
TrolleyMod is a recently conceived and an ongoing project, hence there are many interesting directions for pursuit of its extension. First and foremost in our roadmap is to demonstrate the integration of TrolleyMod with inference techniques such as IRL. Furthermore, our plan for the near-future includes the gradual extension of functionalities to support seamless configuration for more well-known moral dilemmas, and enhance the front and backend to provide better support for handling larger numbers of experiments and volumes of results in a distributed architecture. A similar priority is the development of web and smartphone interfaces to enhance the reach and accessibility of this platform for large scale research. As noted, TrolleyMod is an open-source project with the aim of facilitating research on ethical autonomous vehicles, and is looking forward to contributions and comments from the community.


%
%
%
%
\end{document}